\documentclass[twocolumn,9pt]{article}

\usepackage[margin=1.6cm,top=2cm,bottom=2cm,columnsep=0.5cm]{geometry}
\usepackage{amsmath,amssymb}
\usepackage{graphicx}
\usepackage{xcolor}
\usepackage{booktabs}
\usepackage{microtype}
\usepackage{natbib}
\usepackage[colorlinks=true,linkcolor=blue!70!black,citecolor=blue!70!black,
            urlcolor=blue!70!black]{hyperref}
\usepackage{caption}
\usepackage{authblk}
\usepackage{setspace}
\usepackage{float}

\usepackage[T1]{fontenc}
\usepackage{lmodern}

\usepackage{sfmath}

\newcommand{\lam}{\ensuremath{\hat{\lambda}}}
\newcommand{\Roles}{\textsc{Roles}}
\newcommand{\NoRoles}{\textsc{NoRoles}}

\title{\textbf{Collective AI can amplify tiny perturbations into divergent decisions}}

\author[1,*]{Hajime Shimao}
\author[2]{Warut Khern-am-nuai}
\author[3]{Sung Joo Kim}
\affil[1]{The Pennsylvania State University, Malvern, PA, USA}
\affil[2]{McGill University, Montreal, QC, Canada}
\affil[3]{American University, Washington, DC, USA}
\affil[*]{Corresponding author: hjs5825@psu.edu}

\date{}

\begin{document}
\maketitle

\begin{abstract}
\noindent
Large language models are increasingly deployed not as single assistants but as
committees whose members deliberate and then vote or synthesize a decision.
Such systems are often expected to be more robust than individual models. We
show that iterative multi-LLM deliberation can instead amplify tiny
perturbations into divergent conversational trajectories and different final
decisions. In a fully deterministic self-hosted benchmark, exact reruns are
identical, yet small meaning-preserving changes to the scenario text still
separate over time and often alter the final recommendation. In deployed
black-box API systems, nominally identical committee runs likewise remain
unstable even at temperature 0, where many users expect near-determinism.
Across 12 policy scenarios, these findings indicate that instability in
collective AI is not only a consequence of residual platform-side stochasticity,
but can arise from sensitivity to nearby initial conditions under repeated
interaction itself. Additional deployed experiments show that committee
architecture modulates this instability: role structure, model composition, and
feedback memory can each alter the degree of divergence. Collective AI
therefore faces a stability problem, not only an accuracy problem:
deterministic execution alone does not guarantee predictable or auditable
deliberative outcomes.
\end{abstract}

\section*{Introduction}

Large language models are increasingly being deployed not only as individual
assistants but as committees whose members exchange arguments and then vote or
synthesize a recommendation~\citep{wu2023autogen,chen2023agentverse,li2023camel,hong2023metagpt,park2023generative}.
This architecture is often appealing for a simple reason: groups are expected
to be more robust than individuals. If one model is brittle, idiosyncratic, or
poorly calibrated, deliberation across several agents might seem to offer a
natural way to stabilize judgment. But this intuition need not hold once
interaction becomes iterative. A committee is not merely a collection of
independent votes. It is a feedback system, and feedback systems can amplify as
well as average out small differences. For AI committees intended for policy
advice, institutional decision support, or other high-stakes settings, that
possibility is a basic reliability concern.

The central question of this paper is therefore whether collective AI
deliberation dampens perturbations or amplifies them. Two possibilities must be
distinguished. One is practical: deployed API systems may vary across nominally
identical runs even when users set temperature to 0 and expect near-determinism.
The other is deeper: even if execution is made fully deterministic, nearby
initial conditions such as small meaning-preserving wording changes may still
separate under repeated interaction. This distinction is crucial. If
instability comes only from residual deployment-side randomness, improved
infrastructure or tighter provider controls might largely remove it. But if
exact reruns are identical and nearby initial conditions still diverge, then
the problem lies not only in stochastic reproducibility failure, but in the
dynamics of collective AI deliberation itself.

Prior work has shown that LLM outputs can be sensitive to prompt design and
formatting~\citep{sclar2023quantifying,zheng2023judging}, and long-standing
work in the social sciences has shown that collective judgments depend on
framing, composition, and institutional structure
~\citep{sunstein2002law,tversky1981framing,page2007difference}. Meanwhile,
multi-agent AI systems are increasingly being promoted as tools for
deliberation, coordination, and autonomous decision support. What remains
unclear is whether instability in such systems is mainly a contingent
deployment artifact or a more structural property of iterated interaction among
language-model agents. More broadly, we still lack an empirical stability map
for collective AI: when small perturbations remain local, when they grow, and
which protocol choices amplify or attenuate that growth.

We address this problem by studying five-agent LLM committees as dynamical
systems observed over repeated rounds of deliberation. The paper combines two
complementary empirical regimes. First, we construct a fully deterministic
self-hosted benchmark in which exact reruns are artifact-identical, and
perturb only the initial condition through meaning-preserving variants of the
scenario text. This benchmark provides the cleanest test of whether collective
AI can exhibit sensitive dependence after rerun stochasticity has been removed.
Second, we analyze deployed black-box API committees by holding prompts fixed
and measuring instability across repeated nominal reruns at temperature 0.
Together, these regimes allow us to separate deterministic sensitivity to
nearby initial conditions from practical rerun instability in real-world
deployment. Across both settings, we quantify trajectory separation using an
empirical Lyapunov-style divergence index derived from the evolution of
committee mean preferences over time.

Our main finding is that collective AI can amplify tiny perturbations into
divergent trajectories and different final decisions. In the deterministic
hosted benchmark, exact reruns collapse to a single trace, yet nearby prompt
variants still separate over time and often flip the collective
recommendation. In the deployed API regime, nominally identical committee runs
remain unstable even at temperature 0. Taken together, these results show that
instability in collective AI is not reducible to ordinary sampling noise or
provider-side randomness alone. It can also arise from sensitivity to nearby
initial conditions under repeated interaction itself.

This perspective shifts the evaluation of collective AI from one-shot output
quality to dynamical reliability. A committee may produce a plausible answer in
a single run and still be fragile in the neighborhood of that run. For
governance, the relevant question is therefore not only whether an AI
committee can generate a persuasive recommendation, but whether that
recommendation is stable under perturbations that ought to be innocuous. By
identifying sensitive dependence under deterministic execution and instability
in deployed black-box systems, and by showing that architecture and feedback
memory modulate that instability, this study establishes a foundation for the
next stage of work: linking instability to external decision quality and
designing protocols that reduce divergence without suppressing useful diversity
of reasoning.

\section*{Results}

\subsection*{Collective AI can diverge even under fully deterministic execution}

In plain terms, collective AI can be unstable even when repeated execution is
fully deterministic. In our hosted benchmark, exact reruns of the same
committee pipeline are identical, yet nearby meaning-preserving changes to the
initial scenario text still separate over time and can alter the final
decision. In deployed API systems, nominally identical reruns likewise remain
unstable even at temperature 0. The results below therefore proceed from the
cleaner deterministic test of sensitive dependence to the noisier but
operationally important deployed regime.

\subsection*{Experimental design and instability estimand}

Each run executes a five-agent committee for 20 rounds. At each round, agents
produce both a free-form argument and a structured preference state over the
three policy options. We summarize each run by the committee's mean preference
trajectory on the simplex,
$\overline{\mathbf{p}}^{(r)}_t=\frac{1}{N}\sum_i \mathbf{s}^{(i)}_t$, and
quantify instability by the average pairwise distance between trajectories over
time,
\begin{equation}
D(t)=\frac{2}{R(R-1)}\sum_{i<j}
\left\lVert \overline{\mathbf{p}}^{(i)}_t-\overline{\mathbf{p}}^{(j)}_t \right\rVert_2.
\end{equation}
We then estimate $\lam$ as the slope of $\log D(t)$ over rounds 3--20, so that
positive values indicate exponential separation of trajectories. In the deterministic
hosted benchmark, divergence is computed across nearby meaning-preserving prompt
variants; in the deployed API benchmark, it is computed across repeated
nominal reruns at fixed settings. Unless otherwise noted, target replicate
count is $R=20$ per condition.

\subsection*{Sensitive dependence under deterministic control}

The deterministic hosted benchmark provides the cleanest test of whether
collective AI deliberation can amplify small perturbations after rerun
stochasticity has been removed. We implemented a fully deterministic
self-hosted committee environment using a fixed single-GPU execution path,
greedy decoding, pinned seeds, and deterministic CUDA/PyTorch settings. In
this regime, exact reruns of the same 20-round committee pipeline are
artifact-identical.

We then perturb only the initial condition, using meaning-preserving
surface rephrasings and formatting changes to the scenario text while holding
committee prompts, role prompts, parser structure, voting logic, and all
execution settings fixed. Each hosted condition uses 20 variants: one
canonical wording, 10 AI-generated surface rephrasings, and 9 formatting-only
variants. Candidate variants were manually screened to preserve all numbers,
options, constraints, cases, and the discrete endpoint. Under this design, any
observed divergence reflects sensitivity to nearby initial conditions rather
than stochastic rerun noise.

\begin{figure*}[t]
  \centering
  \includegraphics[width=\textwidth]{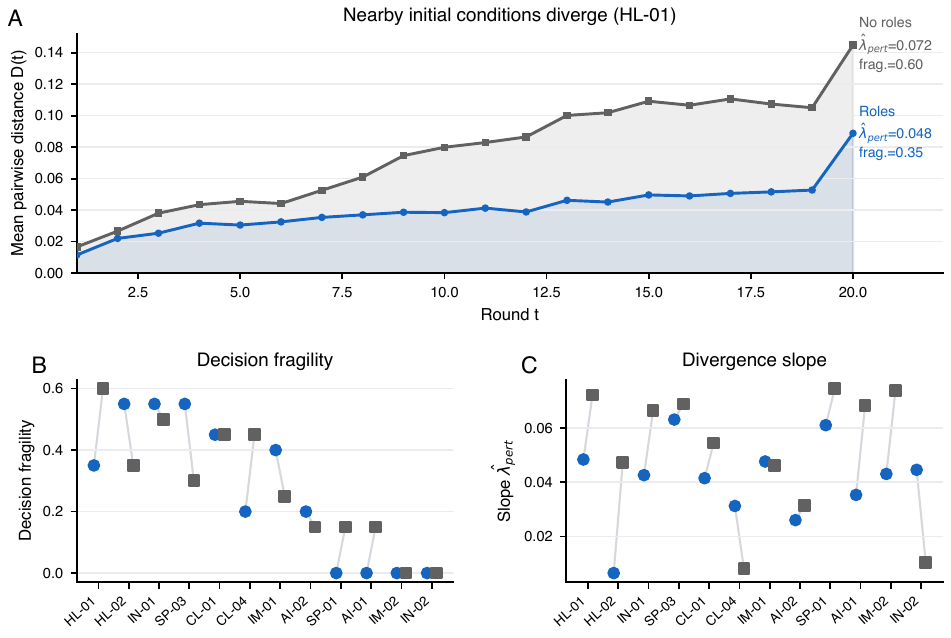}
  \caption{\textbf{Deterministic hosted committees remain sensitive to nearby initial conditions.}
  Hosted runs were executed in a fixed deterministic environment, and exact
  reruns of the same full 20-round committee pipeline were
  artifact-identical.
  (\textbf{A}) Mean pairwise trajectory distance $D(t)$ in HL-01 under
  meaning-preserving perturbations to the scenario text, shown separately for
  \Roles\ and \NoRoles.
  (\textbf{B}) Cross-scenario decision fragility under deterministic
  perturbations, defined as one minus the plurality share across 20 variants.
  (\textbf{C}) Cross-scenario perturbation-divergence slope
  $\hat{\lambda}_{\mathrm{pert}}$.
  The variation shown therefore reflects controlled perturbations of initial
  conditions rather than stochastic reruns. Operationally, a fragility value of
  0.60 means the plurality recommendation appears in only 40\% of nearby
  variants, so the same underlying case can yield materially different
  committee recommendations under tiny wording changes.}
  \label{fig:fig1}
\end{figure*}

Under these controlled perturbations, deterministic committees still fan out
over rounds. Figure~\ref{fig:fig1} shows the resulting mean pairwise distance
curve $D(t)$ for HL-01 and the cross-scenario hosted summaries of perturbation
fragility and divergence. Exact reruns collapse to a single trace, yet nearby prompt variants separate
over time in both \Roles\ and \NoRoles, and for HL-01 the effect appears
within both perturbation families rather than only in the pooled 20-variant
set. Figure~\ref{fig:fig1}B--C shows that this is not a
one-scenario curiosity. Across the 12-scenario hosted benchmark, nearby
variants often change the final decision while producing positive
perturbation-divergence slopes $\hat{\lambda}_{\mathrm{pert}}$. Decision
fragility is especially pronounced in HL-01, IN-01, SP-03, and CL-01, while
divergence slopes remain positive across essentially the full benchmark for
both \Roles\ and \NoRoles. Thus, exact rerun determinism does not guarantee a
robust decision landscape: even semantically conservative perturbations to the
initial scenario can redirect the trajectory of collective deliberation.

In operational terms, the same underlying case can yield materially different
committee recommendations under tiny wording changes that preserve the
underlying facts, options, and endpoint.

Baseline comparisons help interpret this result. When the same deterministic
committees are reduced to one-shot independent judgment, decision fragility
often declines relative to full deliberation, especially in high-fragility
conditions such as HL-01, IN-01, and CL-01. The effect is not perfectly
monotone in every setting, but the average pattern is informative: nearby
prompt variants already contain some latent variation, and iterative feedback
tends to magnify that variation over time. Deliberation is therefore not the
only source of instability, but it is an important amplifier of it.

\subsection*{Practical instability and architectural modulation in deployed API committees}

\begin{figure*}[t]
  \centering
  \includegraphics[width=\textwidth]{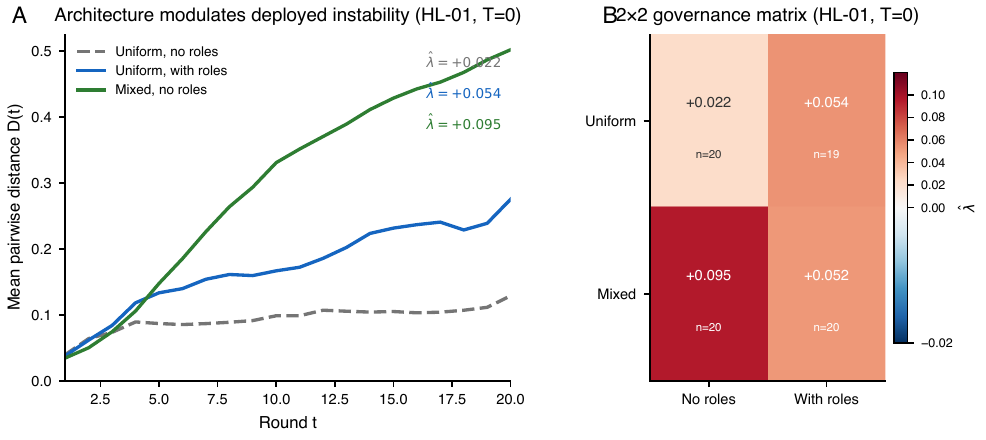}
  \caption{\textbf{Architecture modulates deployed instability (HL-01, $T=0$).}
  (\textbf{A}) Mean pairwise trajectory distance $D(t)$ for three conditions:
  uniform \NoRoles\ (positive but lower-divergence baseline), uniform \Roles,
  and mixed \NoRoles. Adding roles in a homogeneous committee increases
  divergence, and heterogeneous model composition without roles also increases
  divergence.
  (\textbf{B}) 2$\times$2 matrix crossing roles and model composition.
  The mixed+roles cell is less unstable than mixed+no-role, showing non-additive
  interaction. Realized sample sizes are shown in-cell (uniform+roles has
  slight attrition at $n=19$; all other cells are $n=20$).}
  \label{fig:fig2}
\end{figure*}

Figure~\ref{fig:fig2} then provides the operational deployment counterpart to
the hosted benchmark. Figure~\ref{fig:fig2}A shows HL-01 trajectories under
three canonical conditions at $T=0$.
Homogeneous \NoRoles\ is lower-divergence but still positive ($\lam=0.0221$),
while adding role mandates in a homogeneous committee increases divergence
($\lam=0.0541$).
Independently, heterogeneous model composition without roles also yields high
divergence ($\lam=0.0947$). Within the deployed API regime, these results point
to two important architectural amplifiers of instability: role differentiation
in homogeneous committees and model heterogeneity in no-role committees, both
operating on top of a non-zero baseline. These findings show that black-box
committee behavior remains unstable even under nominally fixed settings and is
materially modulated by architecture.

Figure~\ref{fig:fig2}B provides the full HL-01 2$\times$2 matrix.
The mixed+roles cell is substantially less unstable than mixed+no-role
($\lam=0.0519$ vs.\ $0.0947$), indicating non-additivity.
Thus, increasing both diversity dimensions does not monotonically increase
instability~\citep{hong2004groups,page2007difference}.

Supplementary analyses across the 12-scenario benchmark show the same
qualitative ordering:
uniform \NoRoles\ is usually the lowest-divergence configuration, but often
still remains in a positive-$\lam$ regime, while uniform \Roles\ and mixed
\NoRoles\ are more frequently elevated. In other words, nominally identical
reruns remain unstable at temperature 0, and committee architecture can
materially modulate how strongly that rerun variability is amplified in
deployment, but not through any
simple monotone rule such as ``more diversity always means more instability.''
Instead, the instability of deployed collective AI depends on how role
structure, model composition, and interaction protocol are coupled.

\begin{figure*}[t]
  \centering
  \includegraphics[width=\textwidth]{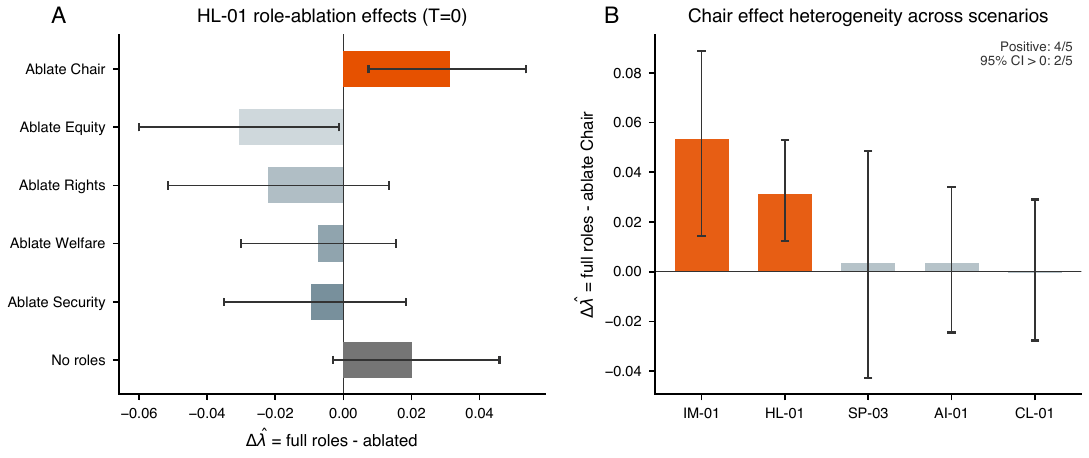}
  \caption{\textbf{Chair channel within role-structured deployed committees.}
  (\textbf{A}) HL-01 role-ablation effects on
  $\Delta\lam=\lam_{\text{full roles}}-\lam_{\text{ablated}}$.
  Chair ablation yields the largest reduction among role mandates
  (with a no-role reference bar shown for context).
  (\textbf{B}) Scenario-level Chair effect across IM-01, HL-01, CL-01, SP-03,
  and AI-01, reported as
  $\Delta\lam=\lam_{\text{full roles}}-\lam_{\text{ablate Chair}}$.
  Point estimates are positive in all five scenarios, with strongest and
  best-resolved effects in IM-01 and HL-01.}
  \label{fig:fig3}
\end{figure*}

Within this deployed/API regime, role structure is therefore one important
architectural channel of amplification, but not the only one. The Chair result
in Figure~\ref{fig:fig3} should be interpreted in this narrower context. In
HL-01, role-ablation effects are reported on the same
$\Delta\lam=\lam_{\text{full roles}}-\lam_{\text{ablated}}$ scale used in the
cross-scenario panel, and Chair ablation yields the largest reduction; other
single-role ablations are near zero or negative in this scenario. Across
five scenarios (IM-01, HL-01, CL-01, SP-03, AI-01), the Chair effect remains
directionally positive, with strongest evidence in IM-01 and HL-01. Thus, when
committees are role-structured, the synthesis-focused Chair channel is often a
strong amplifier, but it is not the sole or universal source of instability in
collective AI; the effect is heterogeneous rather than universal.

The API results therefore make a complementary claim to the hosted benchmark.
They show that instability is not only a feature of carefully controlled
perturbation experiments, but also an operational property of real black-box
systems at nominally fixed settings. Because the baseline noise floor in this
regime is already non-zero, API reruns provide weaker evidence about
growth-from-small-perturbation than the hosted benchmark does. But they provide
stronger evidence about practical unreliability: institutions using deployed
committees should not assume that $T=0$ is enough to make collective behavior
stable or reproducible.

\subsection*{Feedback-memory interventions identify an amplification pathway}

\begin{figure*}[t]
  \centering
  \includegraphics[width=\textwidth]{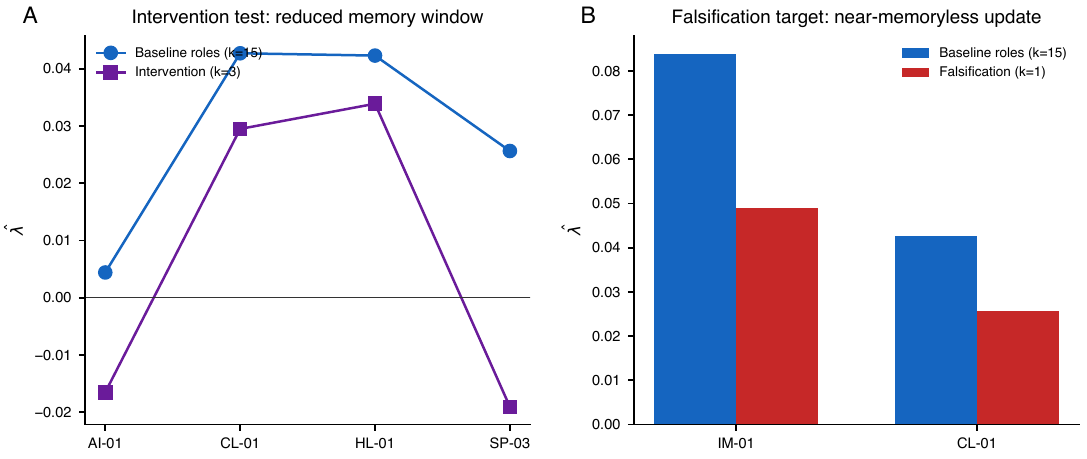}
  \caption{\textbf{Feedback-memory interventions attenuate instability.}
  (\textbf{A}) Intervention: reducing memory window from $k=15$ to $k=3$
  attenuates divergence across four scenarios.
  (\textbf{B}) Falsification target: collapsing memory to $k=1$ lowers $\lam$
  in IM-01 and CL-01 relative to baseline \Roles.
  Together these tests support a feedback-memory amplification pathway.}
  \label{fig:fig4}
\end{figure*}

The deterministic result shifts the interpretation of the intervention tests.
If instability survives under exact rerun determinism, then protocol changes
that reduce amplification are especially important. Figure~\ref{fig:fig4}
reports two such tests. First, reducing argument-memory depth from $k=15$ to
$k=3$ lowers $\lam$ in all four tested scenarios (AI-01, CL-01, HL-01, SP-03).
Second, a stricter memory-collapse variant ($k=1$) in IM-01 and CL-01 lowers
$\lam$ relative to baseline \Roles. These tests do not eliminate divergence in
all settings, but they support a common interpretation across both hosted and
deployed regimes: recursive reuse of prior arguments is one mechanism by which
small perturbations grow into larger trajectory differences over time.
The intervention results therefore move the analysis beyond diagnosis alone and
provide initial evidence that stability in collective AI is at least partly a
protocol-design question.

\subsection*{Interpretation and scope of the instability measure}

Main-text results focus on temperature $T=0$ to isolate structural effects from
sampling-temperature variation.
Complementary analyses in SI show that the deployed-instability signature
persists across a wider temperature range, consistent with a structural rather
than purely thermal interpretation.
The deterministic hosted extension sharpens this point: once exact reruns are
made identical, nearby meaning-preserving perturbations still separate.
Our usage of ``sensitive dependence'' and ``instability'' is therefore
empirical rather than theorem-level: $\lam$ and
$\hat{\lambda}_{\mathrm{pert}}$ quantify growth in trajectory separation under
deployed reruns and deterministic prompt perturbations, respectively; they
should be interpreted as operational instability indices rather than formal
proofs of deterministic chaos in the strict mathematical sense
~\citep{eckmann1985ergodic,ott2002chaos}. That limitation does not weaken the
practical result. The key empirical finding is that small perturbations often
fail to remain local under iterative collective AI deliberation, even when
exact reruns are deterministic. For deployment and governance, that is already
a consequential property.

\section*{Discussion}

Multi-LLM committees are often motivated as a way to make AI decisions more
robust. If one model is noisy, brittle, or idiosyncratic, then deliberation
across several agents might be expected to average out error and yield a more
reliable collective judgment. Our results show that this intuition can fail
once interaction becomes iterative. A deliberating committee is not merely a
collection of independent votes. It is a feedback system, and feedback can
amplify as well as stabilize. Across our experiments, small perturbations were
often enough to redirect collective trajectories and, in some cases, alter the
final decision itself. The central implication is that collective AI has a
stability problem, not only an accuracy problem.

The deterministic hosted benchmark is especially important for interpreting
this phenomenon. In that regime, exact reruns of the same full committee
pipeline were artifact-identical, so the usual explanation of reproducibility
failure---residual stochasticity in execution---was removed by construction.
Yet nearby initial conditions still separated over time when we introduced only
meaning-preserving perturbations to the scenario text, and those perturbations
often changed the final recommendation. This shifts the interpretation of
collective AI instability in a fundamental way. The problem is not only that
deployed black-box systems may remain noisy even when users set temperature to
0. It is also that repeated deliberation can exhibit sensitivity to nearby
initial conditions even under fully deterministic execution. Exact repeatability
of a pipeline, therefore, does not guarantee robustness of its surrounding
decision landscape.

This finding sharpens the practical implications for governance, auditing, and
institutional use. A single deliberation trace may look coherent, well
reasoned, and institutionally plausible while still being locally fragile. In
the deterministic benchmark, some scenarios produced materially different
recommendations under wording changes that preserved the underlying facts,
options, and endpoint. Under such conditions, one observed transcript cannot
safely be treated as representative of the committee's broader behavior. The
relevant object is not only the realized trajectory, but the neighborhood
around that trajectory: whether nearby inputs remain nearby under deliberation,
or instead fan out into different decision basins. For collective AI,
auditability therefore requires more than inspecting one persuasive run. It
requires evaluating stability under perturbations that ought to be innocuous.

The deployed API results add a second layer to this picture. They show that
practical instability remains present in real black-box systems even at
temperature 0, where many users expect near-determinism, and that committee
design modulates how strongly this instability appears. In the deployed
regime, role structure and model composition both affect divergence, but not
in a simple monotone way. This matters because it shows that instability is
not just an unavoidable background nuisance; it is partly architecture-dependent.
The empirical lesson is not that there is a single ``most diverse'' or ``most
structured'' design that uniformly destabilizes committees, but rather that
collective AI behavior depends on how interaction protocols, institutional
roles, and model composition are coupled. That makes stability an architectural
property to be studied, not an incidental by-product to be ignored.

The intervention results provide initial evidence on mechanism as well as
diagnosis. Shortening the memory of prior arguments attenuated divergence
across tested scenarios, and stricter memory-collapse variants further reduced
instability in the settings examined. These manipulations do not eliminate
divergence in all cases, but their value is mechanistic: they suggest that
recursive reuse of prior arguments is one pathway through which perturbations
are amplified over time. In that sense, instability in collective AI is not
fully opaque. It is mediated by identifiable features of deliberation
protocols, especially the degree to which committees repeatedly recycle and
react to their own recent history. This is important because it moves the
paper beyond documenting fragility alone and toward identifying design leverage.

One promising next step is to treat stabilization of collective AI as a control
problem rather than only an evaluation problem. In nonlinear dynamics, the
control-of-chaos literature studies how small, carefully designed perturbations
can stabilize otherwise unstable trajectories without eliminating the
underlying system's flexibility. Our intervention results are much more
preliminary, but they point in a similar direction: instability in multi-LLM
deliberation appears to be shaped by identifiable feedback pathways and
protocol choices, suggesting that targeted interventions may be able to
suppress perturbation amplification without collapsing deliberation into a
trivial one-step vote. An important future agenda, therefore, is to adapt this
broader control perspective to collective AI, asking which minimal
interventions---through memory management, moderator design, bounded
interaction, or adaptive summarization---can improve stability while
preserving the diversity and creativity that make deliberation valuable.

Our claims are empirical and deliberately limited. We do not claim a formal
mathematical proof of deterministic chaos, nor do we claim that all multi-LLM
committees are unstable under all tasks, models, or protocols. Rather, the
contribution of this paper is to show that within a realistic and increasingly
important family of deliberative AI systems, instability is experimentally
inducible, partly controllable, and not eliminated by deterministic execution
alone. That conclusion already challenges a common implicit assumption in
current multi-agent AI practice: that once temperature is lowered and outputs
appear plausible, collective behavior can be treated as effectively stable.
Our results suggest instead that stability must be measured directly.

More broadly, this paper argues for evaluating collective AI as a dynamical
system rather than as a one-shot aggregation device. A committee may produce a
plausible answer in a single run and still be fragile in the neighborhood of
that run. For governance, the relevant question is therefore not only whether
an AI committee can generate a persuasive recommendation, but whether that
recommendation is stable under perturbations that should be innocuous. This
perspective opens a broader research agenda. One next step is to connect
instability to external performance criteria such as accuracy, calibration,
welfare consequences, and decision harm. Another is to develop protocol
designs that reduce perturbation amplification without suppressing useful
diversity of reasoning. Before collective AI can be trusted as a deliberative
technology, its decision landscapes must be shown to be stable, not merely
plausible.

\section*{Materials and Methods}

\paragraph{Protocol.}
We use a windowed-summary deliberation protocol for five-agent committees run
over 20 rounds. At each round, every agent receives the scenario packet, a
committee state table, and a transcript window of prior arguments (default
$k=15$), then returns both free-form argument text and a structured state line.
A deterministic parser extracts preference vectors from that state line, and
one automatic repair pass is allowed after malformed output.

\paragraph{Scenarios.}
The benchmark contains 12 structured policy scenarios spanning immigration,
health, income, climate, speech, and AI governance (two scenarios per domain).
Each scenario presents three discrete policy options and a fixed decision
endpoint.

\paragraph{Models and conditions.}
The paper uses two empirical regimes. The main deterministic hosted benchmark
uses Qwen/Qwen2.5-7B-Instruct for artifact-identical reruns and controlled
initial-condition perturbations. The main deployed/API benchmark uses
GPT-4.1-mini for uniform committees.
Mixed-model committees use a harmonized one-Grok lineup:
Chair (GPT-4.1), Welfare (Claude Sonnet 4.6), Rights (Gemini 2.5 Flash),
Equity (Grok-3-mini), Security (GPT-4.1-mini).
Core deployed conditions cross \NoRoles/\Roles\ with uniform/mixed model
composition at $T=0$.
Additional deployed runs cover role ablation and memory-window variants
($k=3$, $k=1$).

\paragraph{Deterministic hosted extension.}
The deterministic hosted benchmark uses Qwen/Qwen2.5-7B-Instruct in a fixed
single-GPU environment (NVIDIA L4, bfloat16) with greedy decoding,
single-process execution, pinned seeds, and deterministic CUDA/PyTorch
settings. Exact reruns of the same committee condition are artifact-identical
at the full pipeline level. In that hosted benchmark, only the scenario text
is perturbed through 20 manually screened meaning-preserving variants while
committee instructions, parser schema, voting logic, and execution settings are
held fixed.

\paragraph{Inference and uncertainty.}
For each condition, we compute the committee mean trajectory and the mean
pairwise distance curve $D(t)$ over rounds. In the deterministic hosted
benchmark, $D(t)$ is computed across nearby prompt perturbations; in the
deployed/API benchmark, it is computed across repeated nominal reruns at fixed
settings. We estimate $\lam$ from log-linear fits of $D(t)$ over rounds 3--20,
and denote the hosted perturbation analogue by
$\hat{\lambda}_{\mathrm{pert}}$.
Bootstrap confidence intervals are computed by replicate resampling where
reported.
Realized $n$ is reported at panel or cell level where relevant.
Reporting follows reproducibility-oriented practice for empirical ML
benchmarks~\citep{pineau2021improving}.

\paragraph{Implementation.}
Experiments and analyses are implemented in Python 3.11 with NumPy, SciPy,
and Matplotlib~\citep{scipy2020,matplotlib2007}.

\section*{Acknowledgments}
We thank colleagues for feedback on early drafts and experimental design.

\section*{Supplementary Materials}
Supplementary Text S1--S3\\
Supplementary Figures S1--S7\\
Supplementary Tables S1--S8\\
Data S1: JSONL run artifacts\\
Code S1: Experiment and analysis scripts

\clearpage
\onecolumn

\clearpage
\twocolumn
\bibliographystyle{unsrtnat}
\bibliography{refs}

\end{document}